\documentclass[10pt,twocolumn,letterpaper]{article}

\usepackage{cvpr}              

\definecolor{cvprblue}{rgb}{0.21,0.49,0.74}
\usepackage[pagebackref,breaklinks,colorlinks,allcolors=cvprblue]{hyperref}
\usepackage[accsupp]{axessibility}
\usepackage{soul}
\usepackage{algorithm}
\usepackage{algpseudocode}
\usepackage{mathrsfs}
\usepackage{hyperref}
\usepackage{xcolor}
\usepackage{pifont}
\newcommand{\cmark}{\textcolor{green}{\ding{51}}}
\newcommand{\xmark}{\textcolor{red}{\ding{55}}}
\usepackage[margin=1in]{geometry}
\usepackage{bm}
\newcommand{\comment}[1]{}

\def\paperID{*****} 
\def\confName{CVPR}
\def\confYear{2025}
\usepackage{pifont}

\title{Compositional Image-Text Matching and Retrieval by Grounding Entities}

\author{
Madhukar Reddy Vongala,
Saurabh Srivastava,
Jana Košecká \\
Department of Computer Science, \\George Mason University,\\Fairfax, VA, USA \\
\texttt{\{mvongala, ssrivas6, kosecka\}@gmu.edu}
}

\begin{document}
\maketitle
\begin{abstract}
Vision-language pretraining on large datasets of images-text pairs is one of the main building blocks of current Vision-Language Models. 
While with additional training, these models excel in various downstream tasks, including visual question answering, image captioning, and visual commonsense reasoning. However, a notable weakness of pretrained models like CLIP, is their inability to perform entity grounding and compositional image and text matching~\cite{Jiang2024ComCLIP, yang2023amc, Rajabi2023GroundedVSR, learninglocalizeCVPR24}.  
In this work we propose a novel learning-free zero-shot augmentation of CLIP embeddings that has favorable compositional properties. We compute separate embeddings of sub-images of object entities and relations that are localized by the state of the art open vocabulary detectors and  dynamically adjust the baseline global image embedding. 
The resulting embedding is then utilized for similarity computation with text embedding, resulting in a average 1.5\% improvement in image-text matching accuracy on the Visual Genome and SVO Probes datasets~\cite{krishna2017visualgenome, svo}. Notably, the enhanced embeddings demonstrate superior retrieval performance, thus achieving significant gains on the Flickr30K and MS-COCO retrieval benchmarks~\cite{flickr30ke, mscoco}, improving the state-of-the-art Recall@1 by 12\% and 0.4\%, respectively. Our code is available at \url{https://github.com/madhukarreddyvongala/GroundingCLIP}. 
\end{abstract}    
\section{Introduction}
\label{sec:intro}
 Vision-language models (VLMs) have demonstrated significant success in image-text matching and retrieval, achieving strong zero-shot generalization across various vision-language tasks through pretraining on large multi-modal datasets. Models like CLIP \cite{clip} and BLIP-2 \cite{blip2} can effectively retrieve and align images with textual descriptions without requiring task-specific fine-tuning. However, despite their impressive performance, these models face challenges in compositional generalization.
During the pretraining stage, CLIP focuses on aligning images and text as a whole, disregarding the compositional structure of the caption. As a result, its final embeddings often struggle to accurately associate subject and object noun phrases with their corresponding locations in images and face difficulties in handling predicates and relationships in novel configurations~\cite{Rajabi2023GroundedVSR}. Although pretraining involves ingesting large amounts of image-text pairs, it also captures unintended correlations present in the datasets. When certain objects frequently co-occur, the learned representations develop biases, making it difficult to reason about compositional structures that were not seen during training. This often results in errors and hallucinations in downstream VQA tasks~\cite{favero2024multimodalhallucinationcontrolvisual} and poor alignment of image and text embeddings. 
\begin{figure}[t]  
        \centering
\includegraphics[width=\columnwidth]{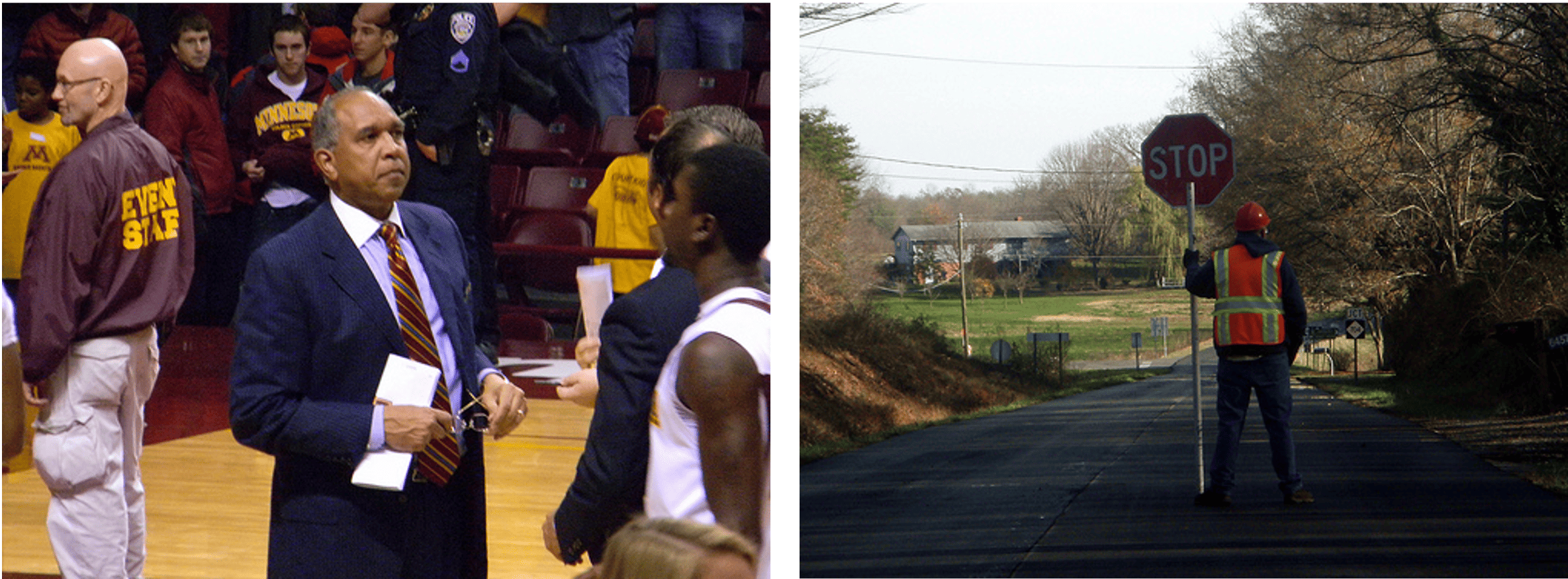} 
        \caption{
        For the caption {\tt A man is holding a sign} (positive image on right side and negative image on left side) Grounding CLIP scores are {\bf 0.2259} and 0.2151 \protect\cmark; CompCLIP are 0.2152 and 0.2226 \xmark, CLIP are 0.2112, 0.2123 \xmark. Note that according to Grounding CLIP the positive images associated with the caption have higher similarity score than negative images.}   
        \label{fig:first}
\end{figure}
For instance, the similarity between the caption ``{\tt a man holding a sign}'' and the image depicting ``{\tt a man holding the glasses}'' (see Figure ~\ref{fig:first}) is high---primarily due to dataset priors rather than compositional computation of similarity.  Similarly, the baseline CLIP embedding of the phrase ``{\tt a man holding a sign}'' will be closer to the image of ``{\tt a man holding the glasses}'', due to a more confident grounding of the subject ``{\tt a man}'' in the right image. This will cause the baseline model to incorrectly rank the similarity of images and text,
affecting the performance of image-text matching and retrieval tasks. 
To address these limitations, we propose Grounding CLIP, a training-free approach that dynamically refines the global image embedding $\bm{I}$ using noun phrase grounding. This enhances the model's ability to better capture the compositional structure of the images and captions, resulting in a more accurate similarity score ${\bf sim}(I,T)$  between image and text.\\

\noindent We utilize the GPT-3.5 language model to break down the text description into object entities and relations, and leverage a state-of-the-art open-vocabulary detector Grounding DINO~\cite{liu2023groundingdino} to localize the phrases corresponding to entities and relations in the image.  
The localized sub-images embeddings are used to dynamically adjust the global image embedding ${\bm I}_c$, which is used in the image-text cosine similarity computation. 
We term the approach Grounding CLIP (GCLIP) and evaluate its performance on several image-text matching and retrieval benchmarks. Our contributions are summarized as follows: \\
\begin{itemize}
  \item We introduce Grounding CLIP, a training free enhancement for compositional image-text matching and image retrieval, leveraging object detection based grounding for improved subject-object alignment.
  \item We achieve state-of-the-art results on ComVG and SVO Probes, consistently outperforming the ComCLIP model and demonstrating better zero-shot compositional understanding using image-text matching evaluation. 
  \item We achieve significant improvements in retrieval tasks on the Flickr30K and MS-COCO datasets using the proposed embeddings improving the state-of-the-art Recall@1 by 12\% and 0.4\%, respectively. 
\end{itemize}

\begin{figure*}
\includegraphics[width=.9\textwidth]{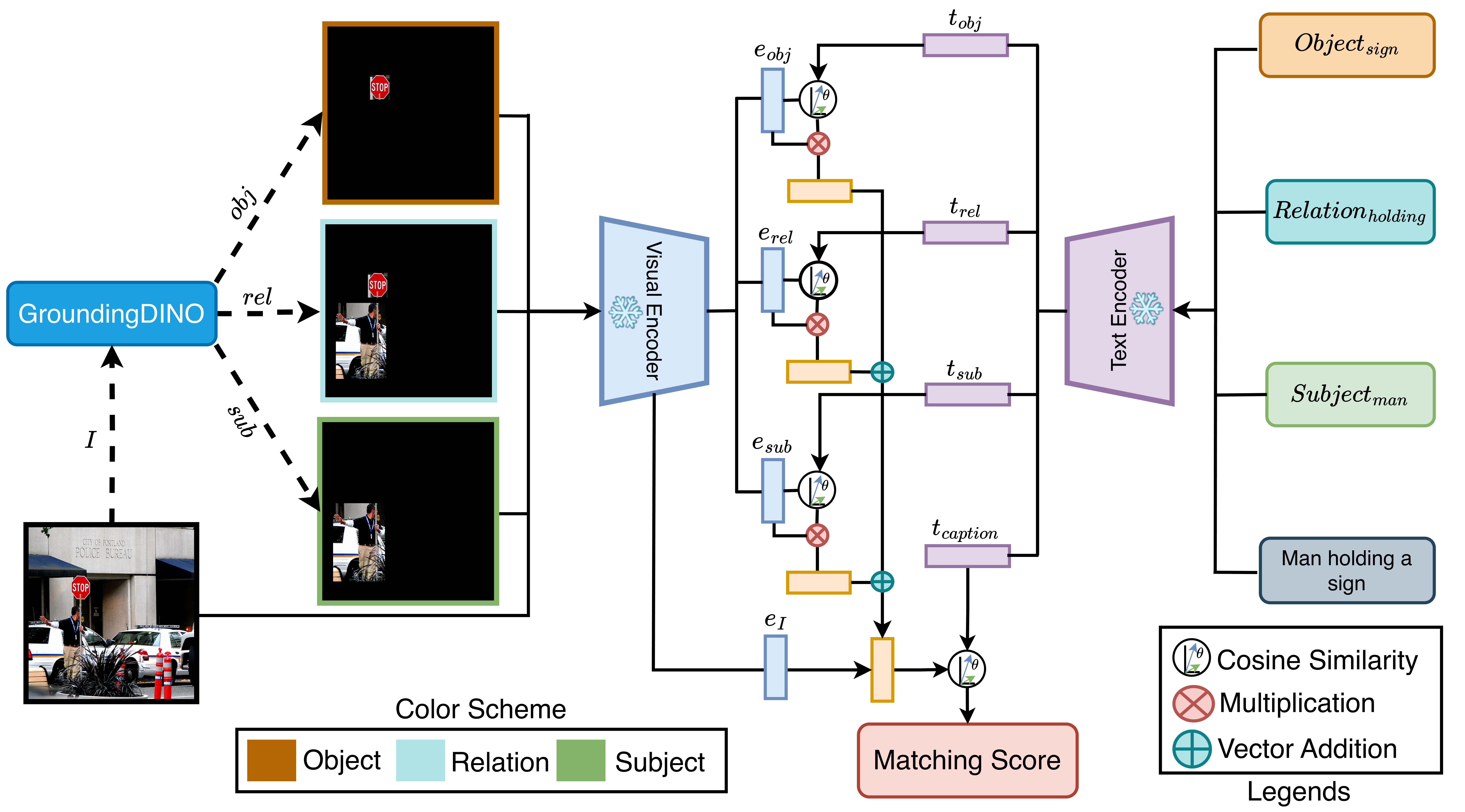} 
        \caption{Overview of the proposed Grounding CLIP (GCLIP) framework. Given a caption such as “A man holding a sign,” we prompt the GroundingDINO with 'man' and 'sign' to localize relevant regions in the image. The resulting sub-images are embedded using CLIP and fused with the global image embedding through a weighted sum based on cosine similarity with their corresponding text phrases. This adjusted image embedding enables more accurate compositional alignment with the caption. }   
        \label{fig:two}
\vspace{-5pt}
\end{figure*}

\section{Related Work}

{\bf Vision Language Models} Foundational Vision and Language Models (VLMs) have demonstrated impressive performance in various vision and language tasks, including visual question answering (VQA), retrieval, image-text matching, referring expression comprehension and  captioning. 
While the earlier models ~\cite{lu2019vilbert,tan2019lxmert,uniter,oscar,vinvl} 
achieved impressive performance using smaller amounts of training data ($\sim$9.7 million image-text pairs) from specially curated datasets, the later generation of VLM's such as ALIGN, BLIP-2, ALBEF, FLAVA, Flamingo, PALI, LLaVA~\cite{align,blip2,albef,flava,flamingo,pali,liu2023llava} relies on the effective pre-training of image and text representations using larger web scale datasets ($\sim$ 400 million image-text pairs) followed by multi-modal learning techniques that combine contrastive learning, masked token modeling, and frozen large language models (LLMs), using another $\sim$ 0.5 million image-text pairs.  These models 
have been shown to improve vision-text alignment, cross-modal reasoning, and zero-shot generalization on variety of datasets and downstream tasks.



\comment{
Many of these models~\cite{align,blip2,albef,flava} start with pretrained image-text representations, such as CLIP~\cite{clip} trained on large web scale dataset of image-text pairs. 
CLIP~\cite{clip} use contrastive learning and a large dataset of image-text pairs ($\sim$400 million) and to align holistic image and text representations, as well as ALIGN \cite{align} (using $\sim$1.8 billion image-text pairs). While CLIP demonstrated high performance on image, scene, and action classification benchmarks ...} 

\paragraph{Probing Studies}
Multiple probing studies of fine-grained understanding capabilities of VLM's have demonstrated poor performance on tasks that require more compositional reasoning. For example, ~\cite{reclip} showed on the CLEVR \cite{clevr} and RefCOCO datasets that if a spatial clause is added to the caption, the performance of image-text matching is at the level of random chance. 
%
Additional probing studies using specially curated datasets demonstrated that these models lack the understanding of attribution, relations, order, verbs, and counting~\cite{stanford-bag-of-words, svo, winoground, kamath2023s}.


\paragraph{Grounding and  Compositionality}
The lack of grounding and compositional generalization of the pretrained representations has been pointed out in several previous works. Authors in~\cite{lewis-etal-2024-clip} focused on CLIP and basic types of compositions (adjectives and nouns, two adjectives and two nouns) in a controlled setting on CLEVR dataset. The linguistic  structure and parsing has been used to do zero-shot referring expression comprehension in~\cite{reclip}; where CLIP was repurposed by region-scoring method that isolates object proposals via cropping and blurring, and passes them to CLIP. This demonstrated the effectiveness of grounding individual noun phrases, relations and combining the evidence using structured probabilistic model. Similar approach was proposed in~\cite{Rajabi2023GroundedVSR} for special case of spatial relations understanding. Poor grounding and localization ability of referring expressions was studied in~\cite{yang2023amc}, where notable improvements were achieved by proposing novel loss function for training multi-modal model with bounding box supervision. 
\comment{
With the success of CLIP and demonstration of its remarkable zero-shot capabilities, several works noted challenges with grounding and localization of the nouns, noun phrases and referring expressions.  Since contrastive learning objective does not include localization objective, the successful grounding typical resulted from existing correlations and co-occurrences in the training datasets. Similar phenomena also lead to hallucinations in visual questions in visual question answering~\cite{favero2024multimodalhallucinationcontrolvisual}. 
}
Additional recent works AMC, OWL, GLIP, Grounding DINO or FLORENCE~\cite{yang2023amc,glip, owl, liu2023groundingdino,xiao2023florence2} used explicit bounding box supervision to improve localization and grounding performance. 
Authors in~\cite{Jiang2024ComCLIP} used GRiT model~\cite{ wu2022gritgenerativeregiontotexttransformer} trained for dense captioning tasks to localize the phrases and dynamically adjusted the global CLIP embeddings to improve similarity score computation. \\ 

\noindent
Our approach is more closely related to~\citet{Jiang2024ComCLIP}, with few notable departures.
We use an LLM for guided decomposition of text into object entities and relations, followed 
by prompting the state-of-the-art open vocabulary grounding model~\cite{liu2023groundingdino} for localization. The use of normalized similarity scores instead of {\tt Softmax} has also been observed as  a contributing factor to the improved performance, due to the fact that individual components of the sentence are not independent.


\section{Grounding CLIP}
\comment{
\begin{figure}[h]  
        \centering
         \includegraphics[width=\columnwidth]{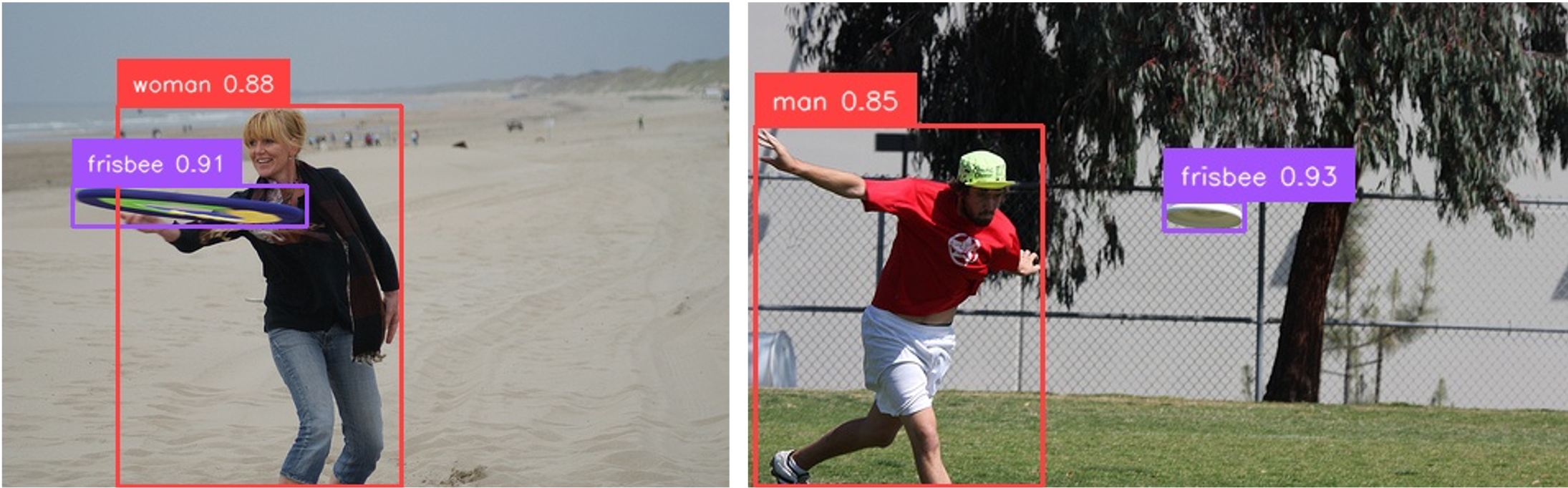}  
        \caption{A woman is throwing a frisbee. GCLIP pos/neg scores 0.2826/0.2581; CompCLIP pos/neg scores 0.2650/0.2728.
        }
        \label{fig:first}
\end{figure}
\begin{figure}[h]  
        \centering
         \includegraphics[width=\columnwidth]{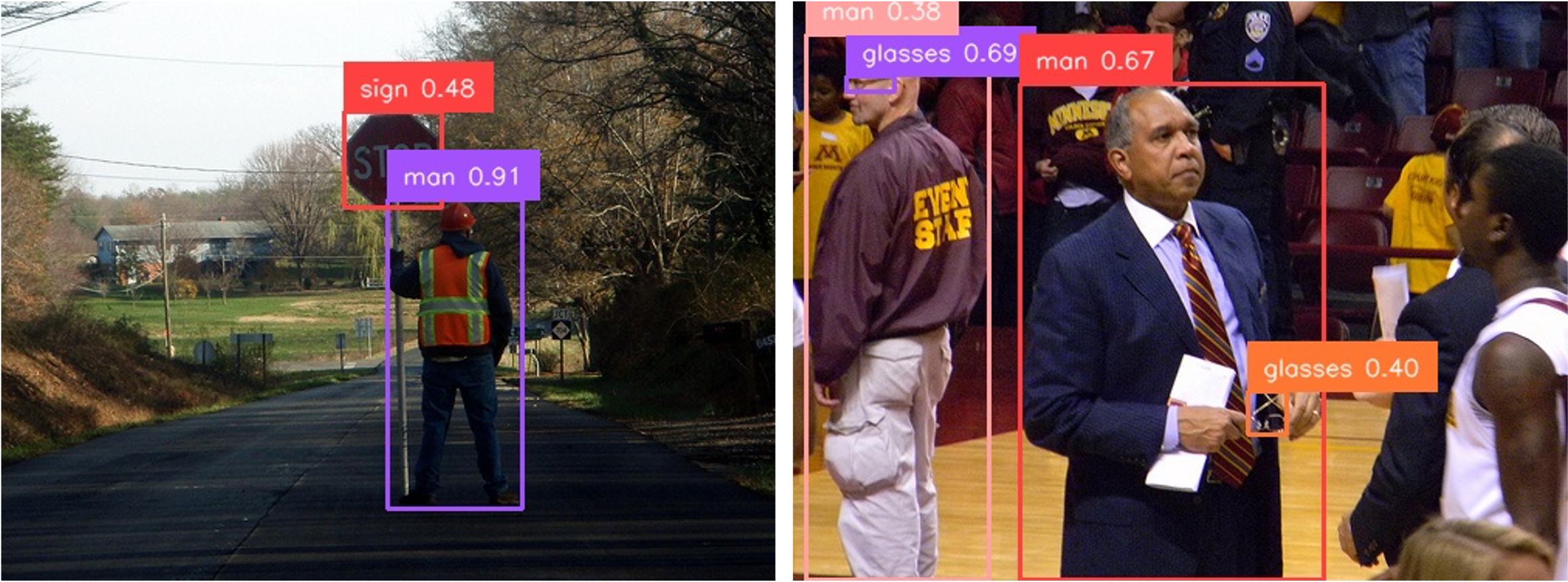}  
        \caption{A man is holding a sign}
        \label{fig:first}
\end{figure}

\begin{figure}[h]  
        \centering
         \includegraphics[width=\columnwidth]{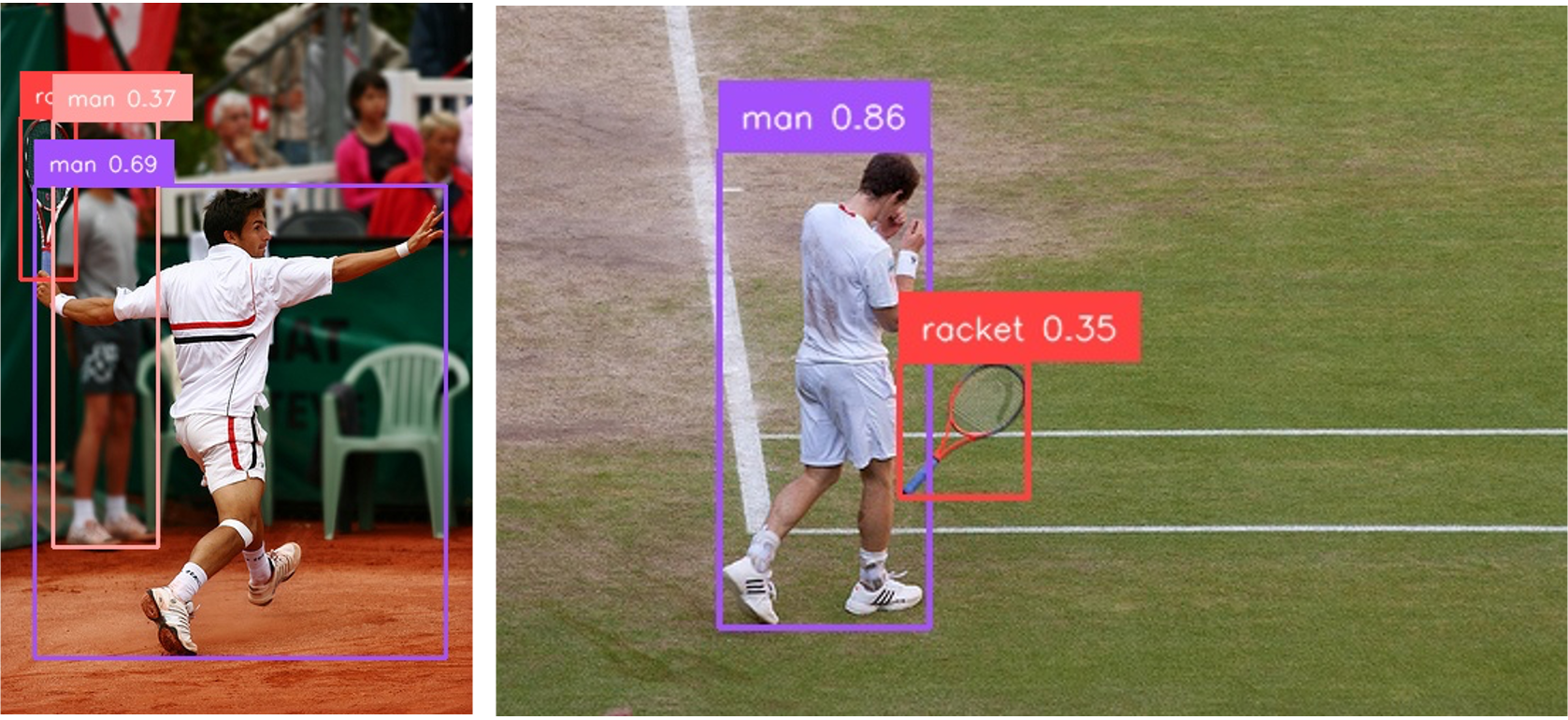}  
        \caption{A man is swinging a racket.}
        \label{fig:first}
\end{figure}

\begin{figure}[h]  
        \centering
         \includegraphics[width=\columnwidth]{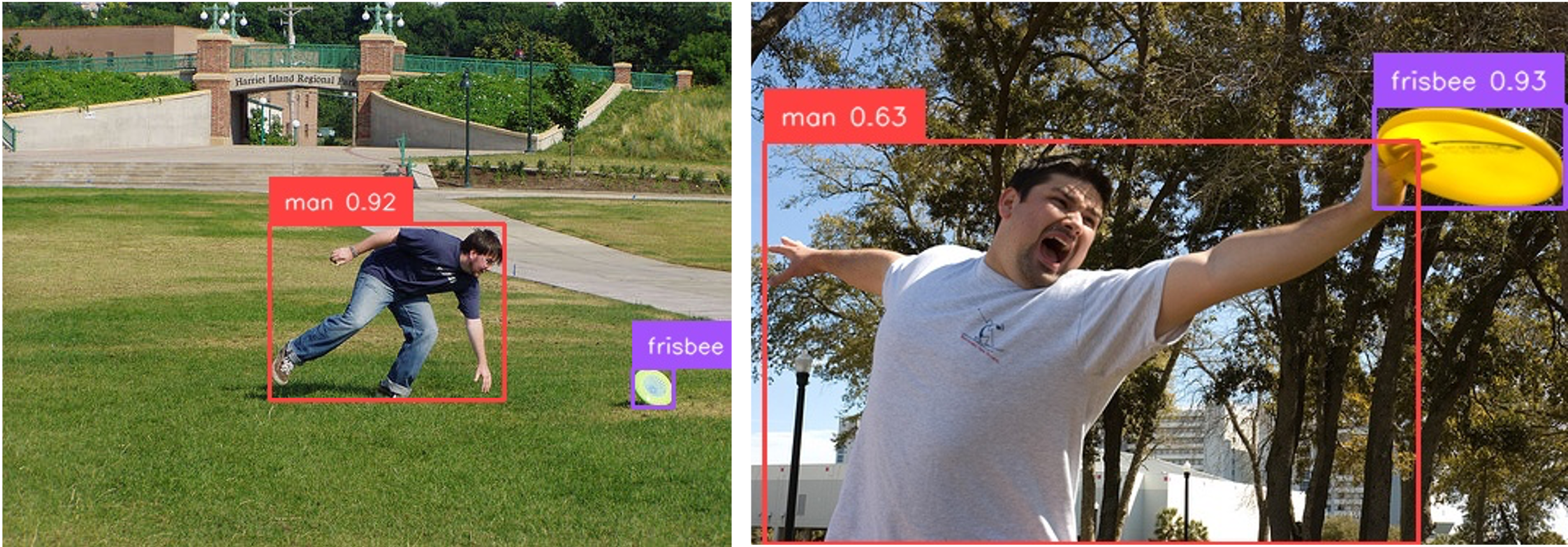}  
        \caption{A man is chasing a frisbee.}
        \label{fig:first}
\end{figure}

\begin{figure}[h]  
        \centering
         \includegraphics[width=\columnwidth]{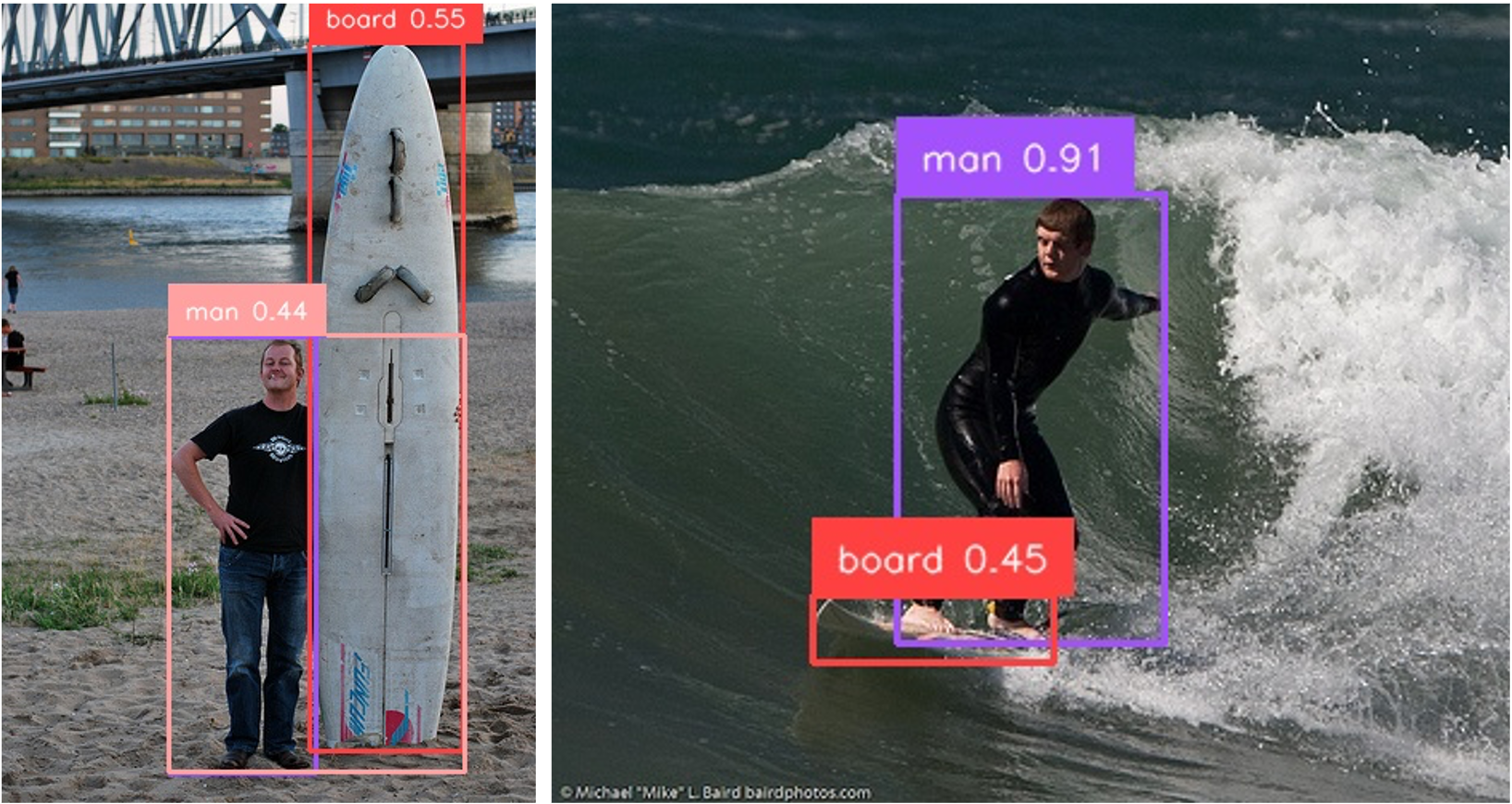}  
        \caption{A man is holding a sign.}
        \label{fig:first}
\end{figure}

\begin{figure}[h]  
        \centering
         \includegraphics[width=\columnwidth]{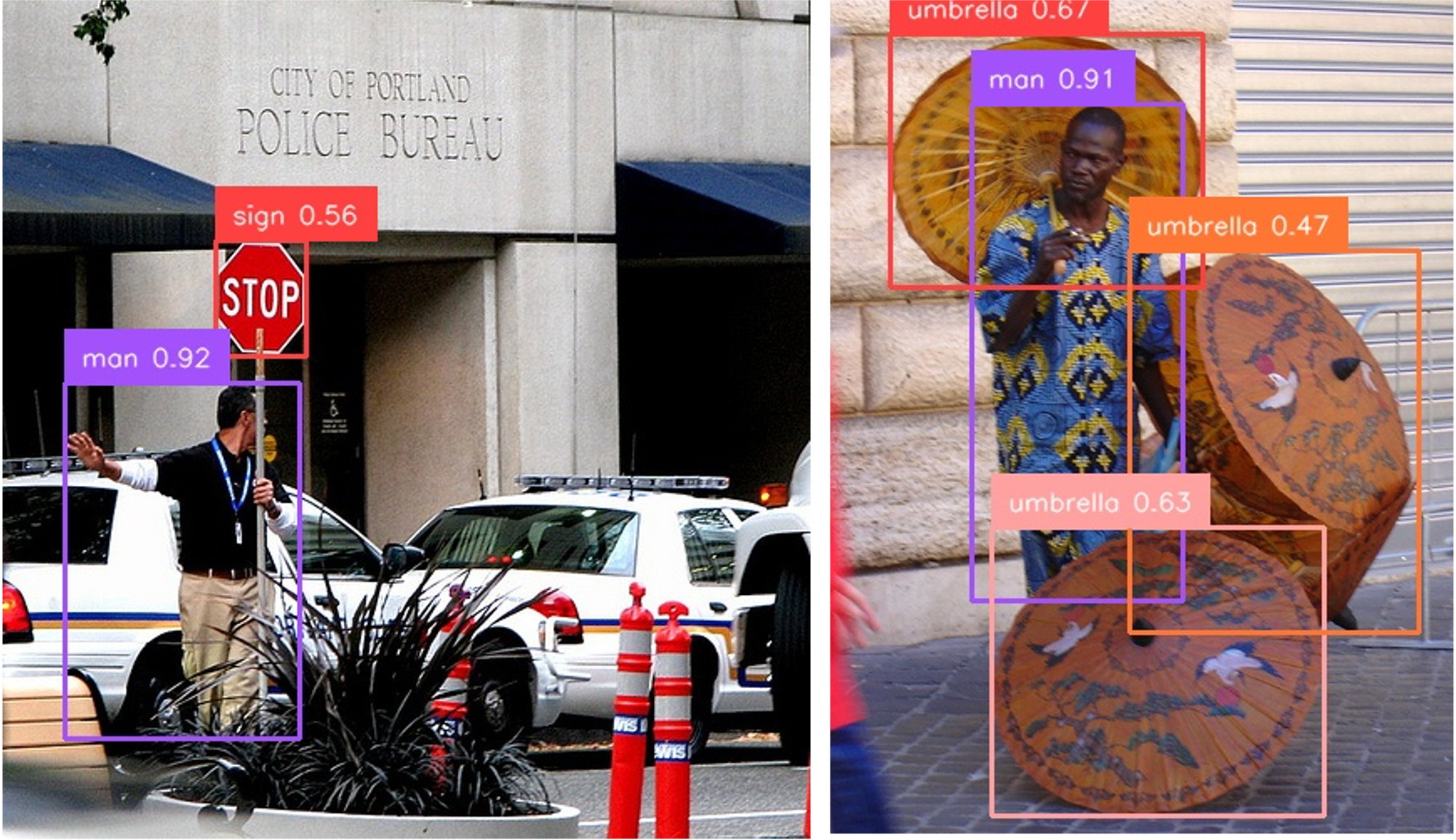}  
        \caption{A man is holding a sign.}
        \label{fig:first}
\end{figure}
}
Our approach, shown in Figure \ref{fig:two}, builds upon \citet{Jiang2024ComCLIP}, that dynamically adjusts image embeddings ${\bm I}_e$ through text query decomposition, to improve grounding and compositionality of image and text alignment. \\

\begin{figure}
\includegraphics[width=\columnwidth]{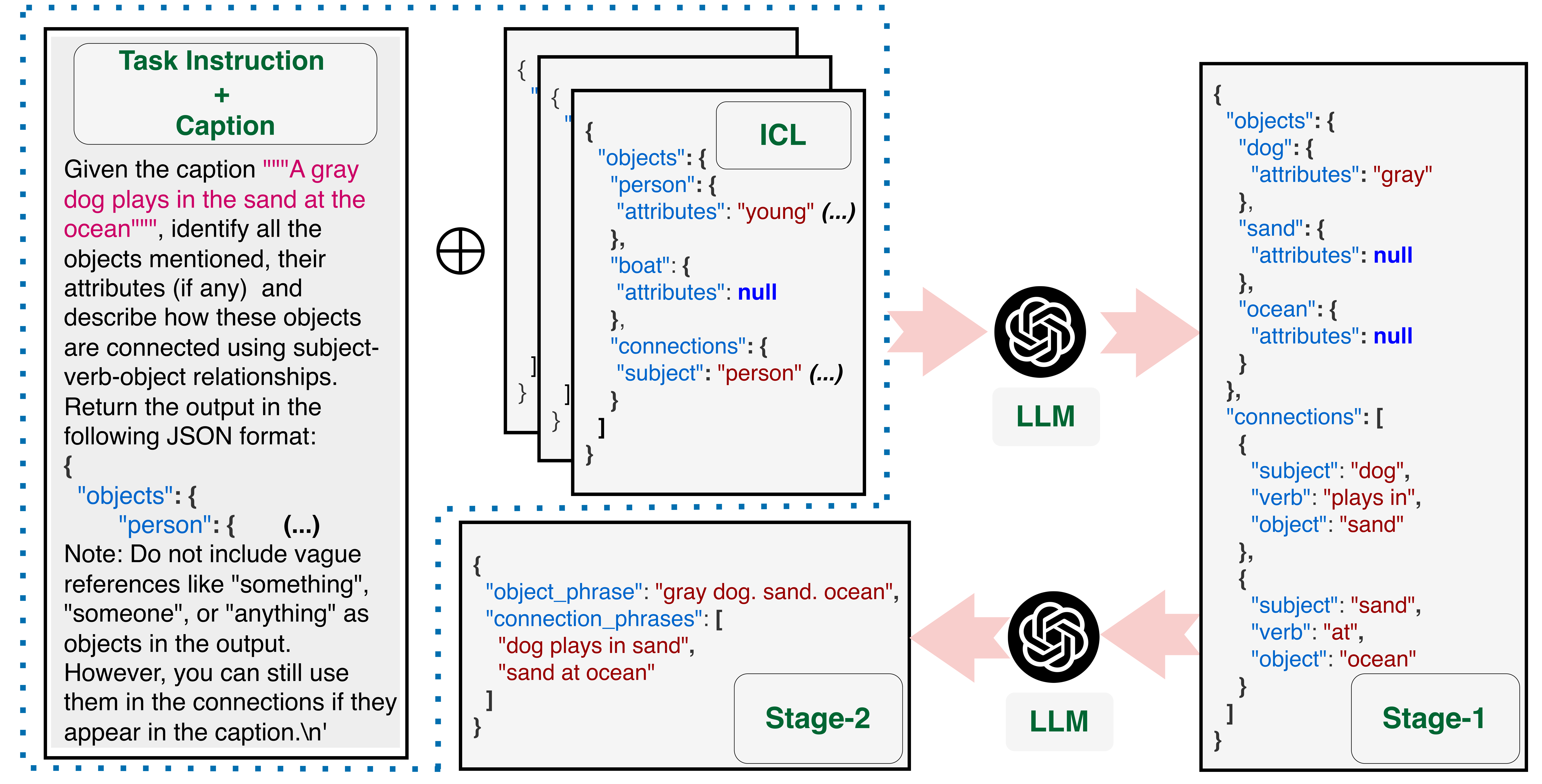} 
        \caption{Two-stage decomposition of the caption “A gray dog plays in the sand at the ocean” using GPT-3.5. In Stage 1, the model identifies objects, attributes, and their relations in structured JSON format. In Stage 2, it generates short natural phrases combining attributes and objects (e.g., “gray dog”, “sand”, “ocean”) and extracts subject-verb-object connections.}   
        \label{fig:stages}
\vspace{-15pt}
\end{figure}

\noindent To address the compositionality 
aspects, we decompose the text into ${(subject, relation, object)}$ tuples or multiple entities and relations ${(ent_1, \hdots, ent_K, rel_1, \hdots, rel_L)}$ using GPT-3.5-turbo. 
First, we prompt the language model to identify all objects in the caption, extract their associated attributes if present and describe how the objects are connected using subject-verb-object relations. We guide the model with detailed descriptions and also with few in-context examples. In the second step, we again prompt the same GPT-3.5-turbo model to generate short natural phrases by combining each object with its attribute as shown in Figure ~\ref{fig:stages}. These phrases are later used as prompts to Grounding DINO~\cite{liu2023groundingdino} 
that localizes the corresponding phrases in the image. 
For SVO probes~\cite{svo} and ComVG datasets~\cite{Jiang2024ComCLIP}, which are used to test the image-text matching, the image captions have simpler triplet structure of the form subject-relationship-object. For more complex captions as encountered in 
Flickr30K \cite{flickr30k} and MS-COCO \cite{mscoco} captions, GPT-3.5-turbo returns $K$ phrases corresponding to $K$ entities present in the
captions and $L$ relations.\\

\begin{figure}[b]  
        \centering
         \includegraphics[width=\columnwidth]{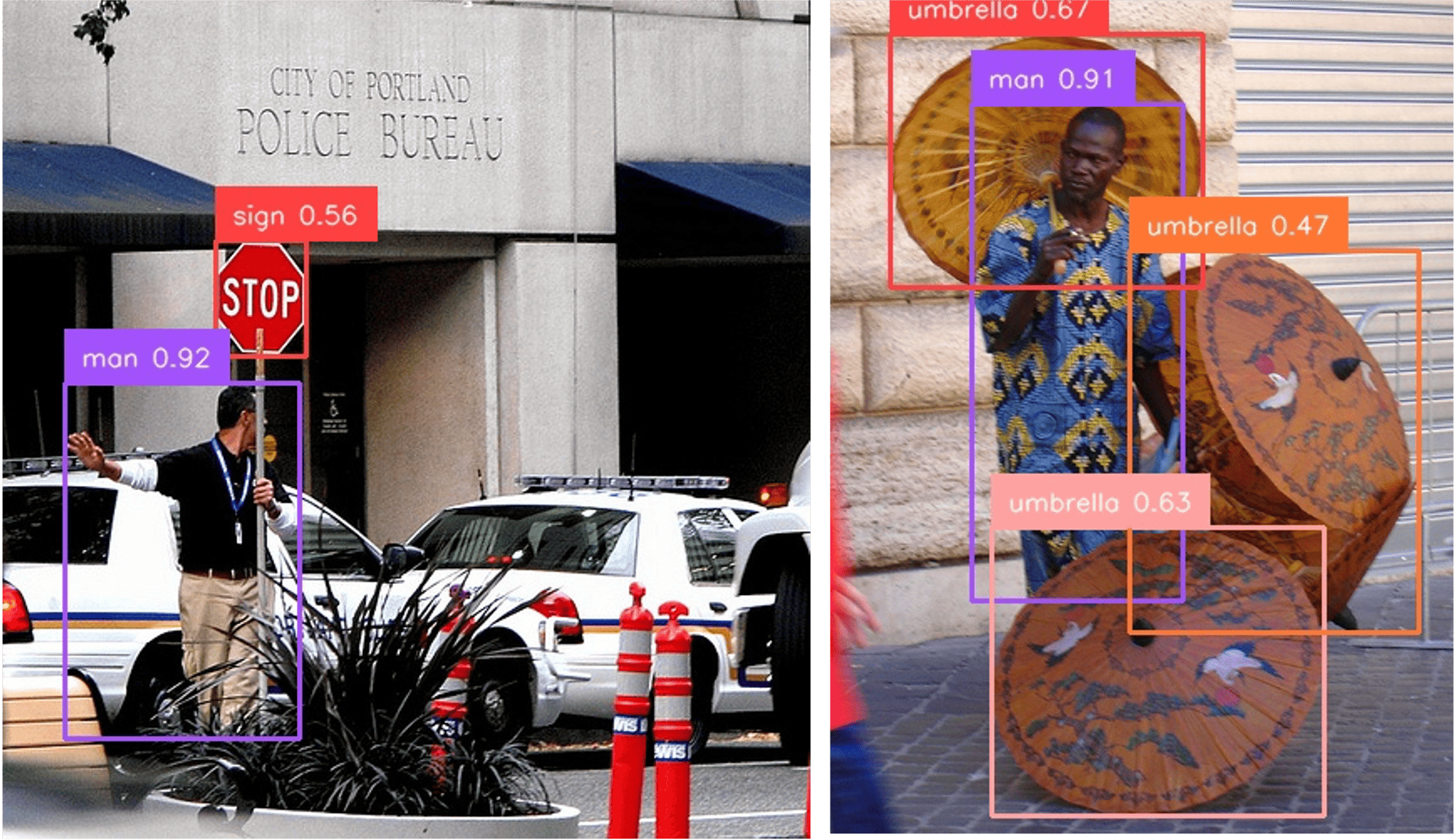}  
        \caption{Examples of bounding boxes returned by Grounding DINO, when prompted by {\tt man} and {\tt sign} on the left and {\tt man} and {\tt umbrella} on the right }
        \label{fig:three}
\end{figure}

\begin{figure}[b]  
        \centering
         \includegraphics[width=\columnwidth]{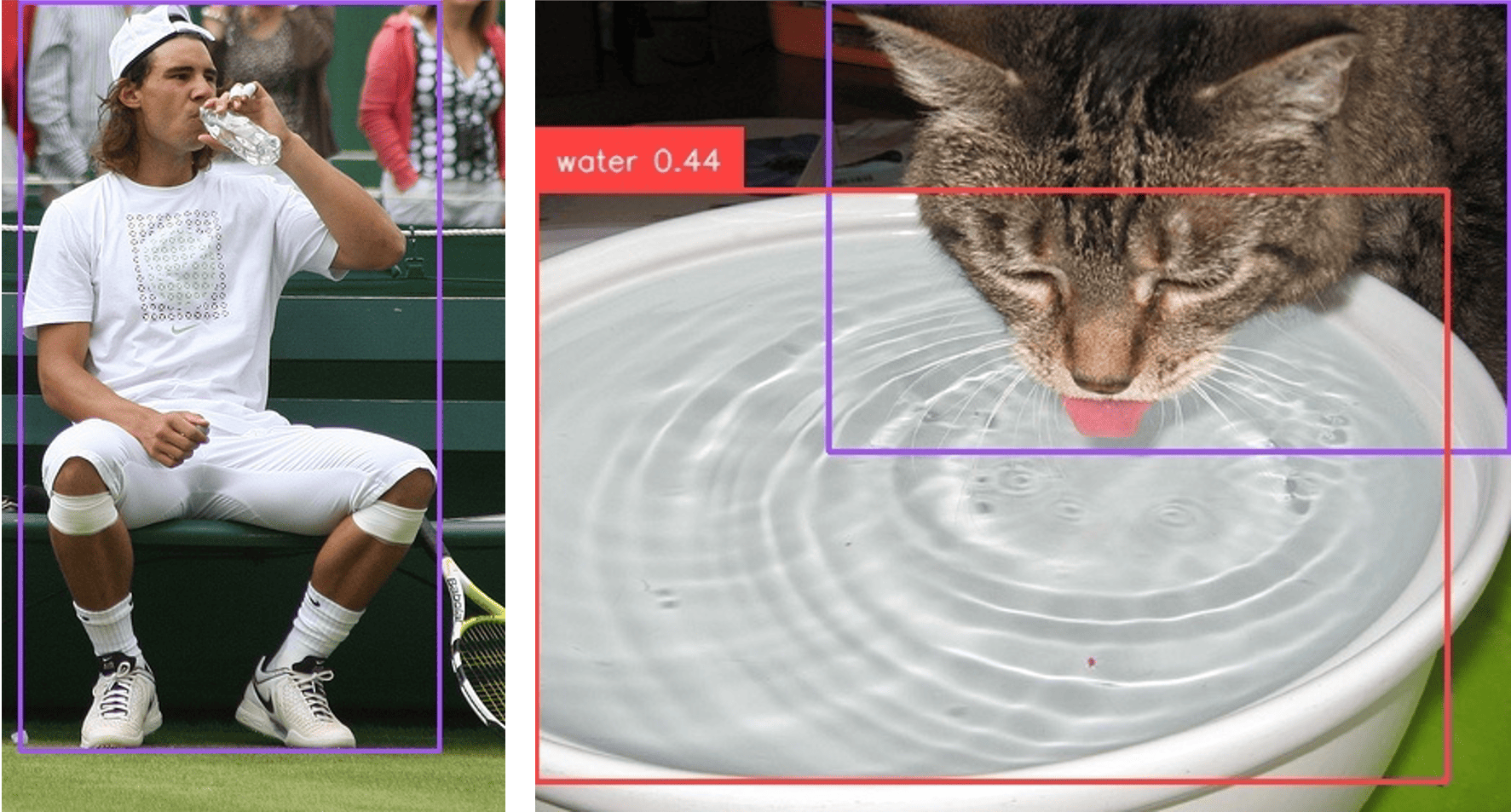}  
        \caption{Examples of bounding boxes returned by Grounding DINO, when prompted by {\tt man}, and {\tt water} on the left and {\tt cat} and {\tt water} on the right.}
        \label{fig:fourth}
\end{figure}

\noindent {\bf Grounding DINO} is a zero-shot open-set object detection model that integrates  DINO (Distillation with No Labels) visual transformer (ViT) trained with self-supervised with language grounding capabilities. The model can detect objects based on arbitrary text queries. It uses contrastive training techniques and cross-modal matching to align textual descriptions with visual features, leading to improved detection of objects, their attributes and referring expressions. It is trained on massive datasets to ensure generalization across various domains (Figure \ref{fig:three} and \ref{fig:fourth}). \\
\begin{algorithm}[t]
    \caption{Compositional Image-Text Alignment}
    \label{alg:compositional_alignment}
    \begin{algorithmic}[1]
        \Require Caption $C$, Image $I$, LLM $\mathcal{M}_L$, Object Detector $\mathcal{M}_O$, Vision-Language Model $\mathcal{M}_V$
        \Ensure Aligned multimodal embeddings

        \State \textbf{1. Text Decomposition}
        \State Query $\mathcal{M}_L$:  
        \Statex \quad $O, A, R \gets \mathcal{M}_L(C)$  
        \Statex \quad $O = \{o_1, o_2, ..., o_n\}$  \Comment{Objects}  
        \Statex \quad $A = \{a_1, a_2, ..., a_n\}$  \Comment{Attributes}  
        \Statex \quad $R = \{(s_1, r_1, o_1), ..., (s_m, r_m, o_m)\}$  \Comment{Relations}  

        \State \textbf{2. Phrase Generation}
        \Statex $P \gets \{ (a_i, o_i) \mid a_i \in A, o_i \in O \}$  
        \Statex $T \gets \{ (s_j, r_j, o_j) \mid (s_j, r_j, o_j) \in R \}$  

        \State \textbf{3. Object Localization}
        \For{$p_i \in P$}
            \State $L_i \gets \mathcal{M}_O(p_i, I)$  \Comment{Detect and localize object}
        \EndFor

        \State \textbf{4. Alignment Computation}
        \State Compute global image embedding: $ \phi_I \gets \mathcal{M}_V(I)$  

        \For{$t_j \in T$}
            \State Extract textual embedding: $ \phi_T \gets \mathcal{M}_V(t_j)$  
            \State Initialize weighted image embedding: $ \phi'_I \gets \phi_I$  

            \For{$p_i \in P$}
                \State Sub-image embedding: $ \phi_{L_i} \gets \mathcal{M}_V(L_i)$  
                \State Alignment score: $s_i = \cos(\phi_{L_i}, \phi_T)$  
                \State Update embedding: $ \phi'_I \gets \phi'_I + s_i \cdot \phi_{L_i}$  
            \EndFor
            
            \State Compute final similarity: $ \mathcal{S}_j = \cos(\phi'_I, \phi_T)$  
        \EndFor

        \State \Return $\{ \mathcal{S}_j \mid t_j \in T \}$  
    \end{algorithmic}
\end{algorithm}
%

\begin{figure}
\includegraphics[width=\columnwidth]{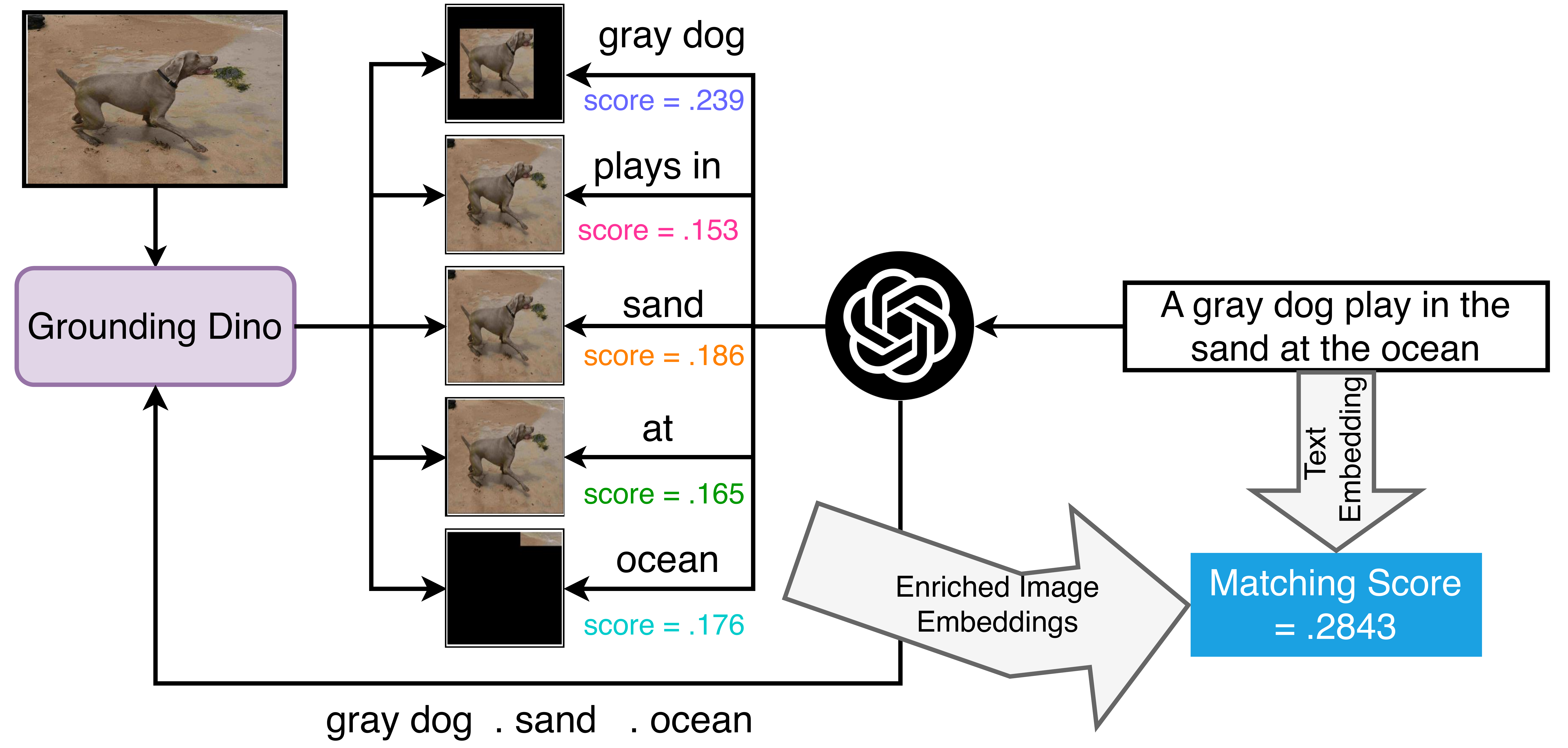} 
\vspace{-5pt}        \caption{Illustration of sub-image generation and compositional fusion in GCLIP. Given the caption from Flickr, “A gray dog plays in the sand at the ocean,” individual noun phrases such as “gray dog,” “sand,” and “ocean” are grounded using GDINO, and corresponding sub-images are extracted. Each sub-image is embedded with CLIP and weighted by its similarity to the associated phrase. These embeddings are combined with the global image embedding to form an enriched representation that improves alignment with the caption.}   
        \label{fig:decompose}
\vspace{-10pt}   
\end{figure}

\noindent
{\bf Sub-image generation}. Given the subject and object noun phrases and their associated bounding boxes localized by Grounding DINO $b_i = (x_i, y_i, w_i, h_i)$ and $b_j = (x_j, y_j, w_j, h_j)$, we mask out the remainder of the image and compute the associated CLIP embeddings ${\bf e}_{subj}$ and ${\bf e}_{obj}$. In case of more complex captions that involve $K$ 
entities, we prompt Grounding DINO with $K$ phrases returned by GPT-3.5-turbo, obtaining $K$ bounding boxes and their associated CLIP sub-image embeddings ${\bm e}_k$. The sub-images associated with relations are determined by bounds of bounding boxes of entities associated with relations. See Figure~\ref{fig:decompose} for more detail. \\ 

\noindent
{\bf Similarity Computation.} We use the normalized form of cosine similarity between constituent sub-images and phrase embeddings 
${\bm e}_i$ and ${\bm t}_j$, i.e., 
\begin{equation}
{\bf sim}({\bm e}_i,{\bm t}_j) = \frac{{\bm e}_i^T {\bm t}_j}{\|{\bm e}_i\| \|{\bm t}_j\|} 
\end{equation}
For ComVG and SVO image-text matching benchmarks, the image captions are made of triples ${(subject,~relation,~object)}$. Corresponding similarity scores with textual embeddings of the associated noun phrases ${\bf t}_{subj}$ and $ {\bf t}_{obj}$ are obtained as: 
\begin{equation}
s_{obj}= \mbox{sim}({\bm e}_{obj}, {\bm t}_{obj}) \: \: \mbox{and} \:  s_{subj}=\mbox{sim}({\bm e}_{subj}, {\bf t}_{subj})
\end{equation}

For relation grounding, we compute embedding of 
sub-image bounded by the extreme coordinates of $b_{obj}$ and $b_{subj}$ and compute its similarity score with the relation, i.e., 
\begin{equation}
s_{rel}= \mbox{sim}({\bm e}_{rel}, {\bm t}_{rel})
\end{equation}
The dynamically adjusted embedding of the entire image ${\bm I}_c$ will be:
\begin{equation}
{\bm I}_c = {\bm e}_I + s_{obj} {\bm e}_{obj} + s_{subj} {\bm e}_{sub} + s_{rel} {\bm e}_{rel}
\end{equation}
where ${\bm e}_I$ is the baseline CLIP embedding of the entire image. 
\noindent
In case of more complex captions that involve $k$ 
entities corresponding to $k$ noun phrases returned by LLM and associated similarity scores $s_j= \mbox{sim}({\bm e}_{j}, {\bm t_j})$ for all $j=1 \hdots k$, the new image embedding is given as 
\begin{equation}
{\bm I}_c = {\bm e}_I + \sum_{k=1}^K s_k {\bm e}_{k} 
\end{equation}
The final similarity score will be $\mbox{sim}({\bm I}_c, {\bm T})$, where ${\bm T}$ is the text embedding of the entire caption. \\\\
%
We found the score $s_k$ computed using normalized embeddings to be more effective than unnormalized scores followed by {\tt Softmax}  normalization used in~\citet{Jiang2024ComCLIP} that assumes the scores are independent. 
\comment{
\begin{figure*}[h]
\begin{minipage}{0.54\columnwidth}
        \centering
         \includegraphics[width=\textwidth]{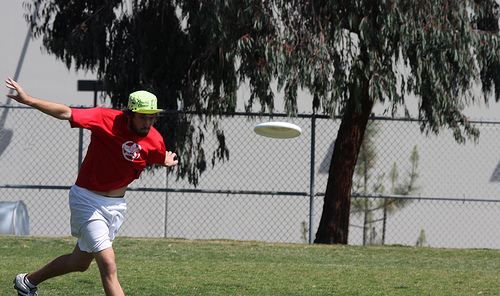}  
        \label{fig:first}
    \end{minipage}
    \begin{minipage}{0.48\columnwidth}
        \centering
        \includegraphics[width=\textwidth]{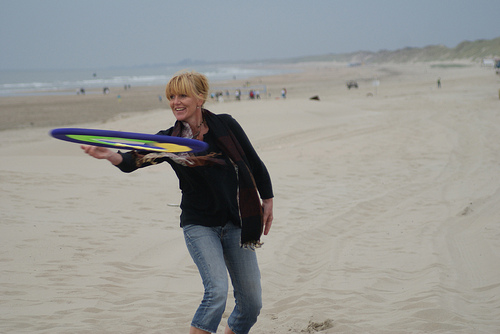}
        \label{fig:second}
    \end{minipage}
    \label{fig:sent_res}
\end{figure*}
\begin{figure*}[h]  
        \centering
         \includegraphics[width=\columnwidthwidth]{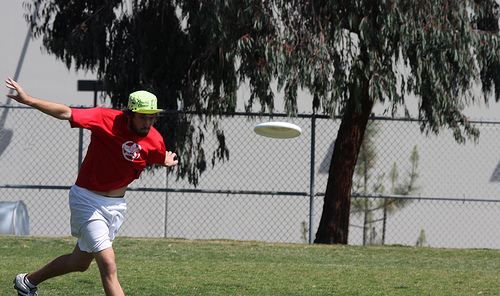}  
        \label{fig:first}
\end{figure*}
}
In summary, the overall process entails breaking down the input image into entity-specific sub-images, computing similarity scores between these sub-images and their corresponding textual representations, and then combining these weighted embeddings with the global image embedding. \\

\noindent CompCLIP~\cite{Jiang2024ComCLIP} achieves
compositional matching through a two stage process, where a dense captioning model (GRiT)~\cite{wu2022gritgenerativeregiontotexttransformer} generates region descriptions, and then a language model (GPT-3.5) filters the captions that best corresponds to the entities in the image. While this approach improves compositionality, it raises two key challenges. First, relying on automated captioning and text heuristics introduces potential errors, due to incomplete or misleading captions that affect the alignment. Second, dense captioning is prone to generating numerous captions for a single subject/object, creating redundant sub-images that increase computational cost and induce inconsistency in similarity scoring. Such challenges make ComCLIP computationally expensive and less interpretable, thus limiting its efficiency in real-world scenarios where precise and effective compositional reasoning is critical. \\

\begin{figure}[h]  
         \includegraphics[width=\columnwidth]{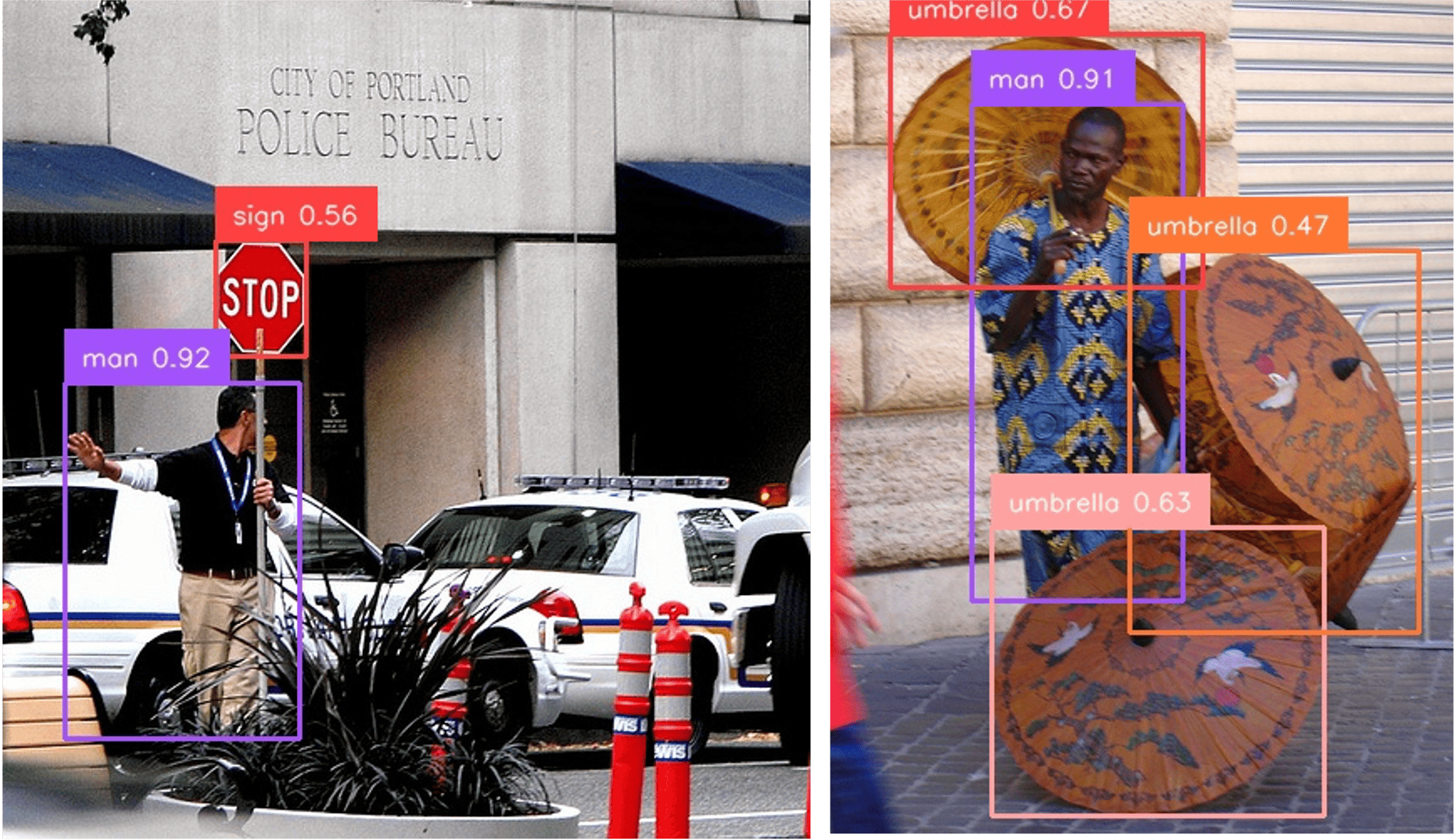} 
          \caption{Positive and negative image for caption {\tt a man is holding a sign}. Positive and negative scores for GCLIP are  0.2052/0.2069 \xmark  and ComCLIP are 0.2140/0.2080 
          \protect\cmark}
        \label{fig:holding}
\end{figure}
\comment{
\begin{figure}[h]  
         \includegraphics[width=\columnwidth]{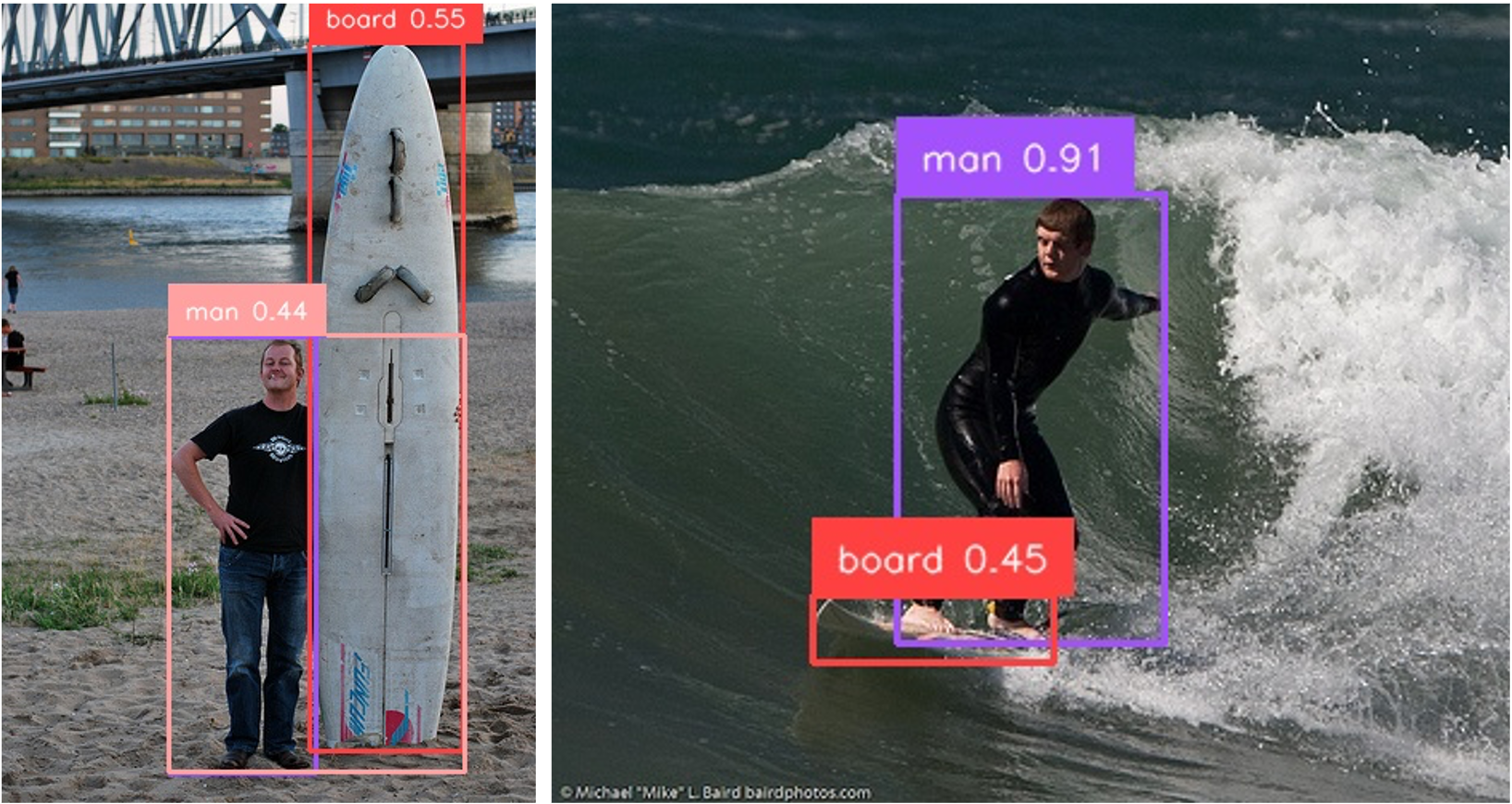} 
         \caption{ Positive and negative image for caption {\tt a man is riding a board}. Positive and negative score for  ComCLIP HIT are 0.2474, 0.2526 and GCLIP MISS are 0.2511, 0.2369}
        \label{fig:riding}
\end{figure}
}
\begin{figure}[h]  
         \includegraphics[width=\columnwidth]{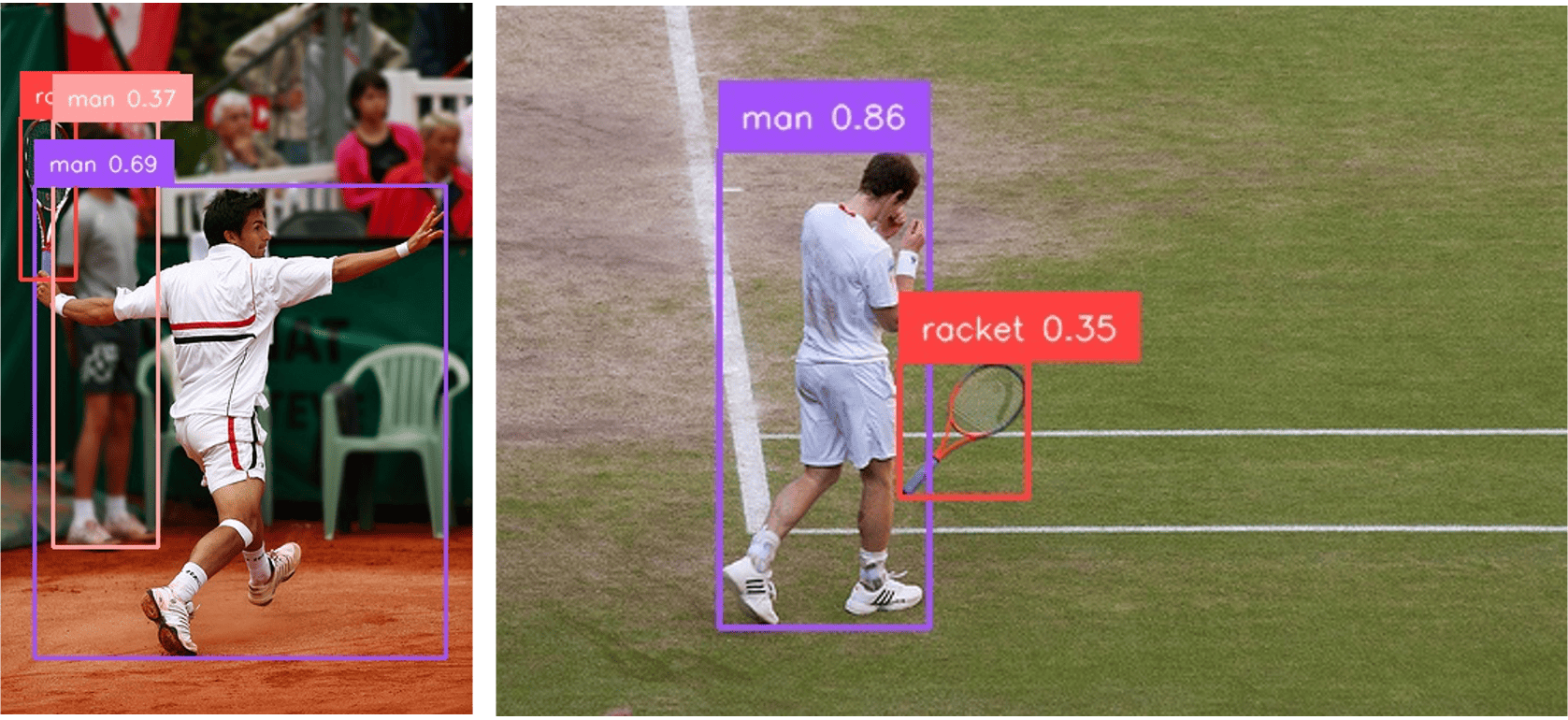} 
          \caption{Positive and negative image for caption {\tt a man is swinging a racket}, negative image differs in verb; Positive/negative scores for GCLIP are  0.2260/0.2101 \protect\cmark and ComCLIP are  0.2153/0.2182. \xmark}
        \label{fig:swinging}
\vspace{-10pt}
\end{figure}

\section{Experiments}
To analyze fine grained compositional image embeddings,  
we evaluate our approach on selected image-text matching  and retrieval benchmarks. \\

\noindent \textbf{Compositional Visual Genome (ComVG)} \cite{Jiang2024ComCLIP} dataset is derived from Visual Genome \cite{visualgenome} and focuses on subject-verb-object (SVO) triplets. This dataset comprises of 524 distinct images, each varying by either the subject, verb, or object. Across these images, there are 5400 relations capturing fine grained variations in SVO combinations. This dataset focuses on understanding structured triplets rather than natural captions. We evaluate our approach on the entire dataset. \\

\noindent \textbf{SVO Probes \cite{svo}} dataset is constructed from Conceptual Captions \cite{sharma2018conceptual} and is designed to evaluate a model's sensitivity to changes in the compositional structure of subject-verb-object triplets. While the dataset contains 36,842 samples, we evaluate our approach on a filtered subset of 13,767 samples, as many of the original images are either unreadable or no longer accessible. \\

\noindent \textbf{Flickr30K \cite{flickr30k}} is a widely used image-text retrieval benchmark containing 31,783 images each annotated with 5 human written captions. The captions are natural, free form descriptions that capture objects, actions and interactions in the scene. Following the evaluation protocol in \cite{Jiang2024ComCLIP}, we randomly select one caption for the test set that has 1000 images. First, CLIP \cite{clip} is used to retrieve the top 10 similar images for each caption. For these top ranked candidates, we apply our Grounding CLIP (GCLIP) approach. \\

\noindent \textbf{MS-COCO \cite{mscoco}} is another large benchmark for image-text retrieval with over 120,000 images paired with five human written captions. Similar to Flickr30K, the captions in MS-COCO describe objects, actions and spatial relationships in the scene using natural language. We follow the same evaluation protocol similar to Flickr30K. Out of 40,000 images from the validation set, we randomly select 1000 images and for each image we select one random caption out of the five captions. CLIP is used to retrieve the top 10 similar images for each caption from this validation set. Then we apply Grounding CLIP (GCLIP) on the top 10 candidates.  \\

\noindent ComVG and SVO probes are used to evaluate image-text matching as they consist of structured subject-verb-object triplets. Flickr30K and MS-COCO are used to assess image-to-text retrieval capabilities based on natural language descriptions.

\begin{table}[h]
    \centering
    \resizebox{\columnwidth}{!}{%
    \begin{tabular}{lcccc}
        \toprule
        \textbf{Vision Encoder} & \textbf{CLIP} & \textbf{OpenCLIP} & \textbf{ComCLIP}$^*$ & \textbf{GCLIP}$_{\text{ours}}$ \\
        \midrule
        ResNet-50 & 82.25 & 82.21 & 83.26 & \textbf{84.79} \\
        ViT-B-32  & 82.45 & 82.41 & 83.12 & \textbf{85.61} \\
        ViT-L-14  & 86.38 & 86.38 & 86.93 & \textbf{87.85} \\
        \bottomrule
    \end{tabular}%
    }
    
    \caption{Image-text matching accuracy \% of different models on the Compositional Visual Genome dataset.($^*$\textit{The numbers are reproduced by us.})}
    \label{tab:comvg_results_1}
\end{table}
\vspace{-10pt}
\begin{table}[h]
    \centering
    \resizebox{\columnwidth}{!}{%
    \begin{tabular}{lcccc}
        \toprule
        \textbf{Vision Encoder} & \textbf{CLIP} & \textbf{OpenCLIP} & \textbf{ComCLIP$^*$} & \textbf{GCLIP}$_{\text{ours}}$ \\
        \midrule
        ResNet-50 & 81.21 & 81.56 & 82.06 & \textbf{82.67} \\
        ViT-B-32  & 81.96 &  80.67& 82.68 & \textbf{83.88} \\
        ViT-L-14  & 84.79 & 85.23 & 85.32 & \textbf{86.29} \\
        \bottomrule
    \end{tabular}%
    }
    \caption{Image-text matching accuracy \% of different models on the SVO-Probes dataset. ($^*$\textit{The numbers are reproduced by us.})}
    \label{tab:svo_results_1}
\vspace{-15pt}
\end{table}
   
\subsection{Baselines}
We compare our approach against several strong vision-language baselines to evaluate improvements in both compositional understanding and retrieval performance. \\

\noindent \textbf{CLIP \cite{clip}:} CLIP is a widely used vision language model trained with contrastive learning on 400M image-text pairs. It is our primary baseline for both image-text matching and retrieval tasks. \\

\noindent \textbf{OpenCLIP \cite{OpenCLIP_2021}:} An open source implementation of CLIP with similar architecture and training objectives. We include it to assess consistency across CLIP variants.  \\

\noindent \textbf{ComCLIP \cite{Jiang2024ComCLIP}:} A training free method that enhances CLIP for compositional reasoning by leveraging dense captioning and localization. This is the most closely related prior work and serves as a key comparison for compositional image-text matching. \\

\noindent \textbf{SLIP \cite{slip} and ComSLIP \cite{Jiang2024ComCLIP}:} SLIP extends contrastive vision-language training with self-supervised objectives. ComSLIP integrates compositional information into SLIP embeddings similar to ComCLIP. \\

\noindent \textbf{BLIP \cite{blip}  and BLIP2 \cite{blip2}:} These models improve multimodal alignment and reasoning by combining Vision Encoder with Large Language Models. We also compare with their compositionally enhanced variants ComBLIP and ComBLIP2.


\subsection{Image-Text Matching}
\label{sec:image_text_matching}
In image-text matching, for each triplet, we localize the subject and object nouns in the image using Grounding DINO. Instead of cropping these regions, we create masked sub-images by blacking out all areas of the image except the detected bounding box for the target entity. This ensures that each sub-image has same dimension as the original image while separating the visual signal associated with a specific entity. To create a relation sub-image, we combine the related subject and object sub-images. This combined sub-image effectively captures the context and interaction between the subject and object. \\

\noindent The original image and its corresponding caption along with the masked sub-images and their associated words (subject, object, relation) are processed through the CLIP's vision and text encoders. We then calculate the cosine similarity between each sub-image embedding and its respective word embedding. These similarity scores are used as weights to compute a weighted sum of the sub-image embeddings. This weighted sum is added to the global embedding of the original image to produce an adjusted image representation ${\bm I}_c$. Finally, we compute the cosine similarity between ${\bm I}_c$ and the embedding of the entire caption. The methodology remains same for the other baselines as well, termed as GSLIP, GBLIP and GBLIP2. \\

\begin{table}[h]
    \centering
    \begin{tabular}{lcccc}
        \toprule
        \textbf{Method} & \textbf{Sub} & \textbf{Rel} & \textbf{Obj} & \textbf{Total} \\
        \midrule
        SLIP & 86.2 & 61.33 & 85.81 & 80.13 \\
        ComSLIP$^*$ & 86.11 & \textbf{63.05} & 89.51 &  82.27\\
        \textbf{GSLIP}$_{\text{ours}}$ & \textbf{87.17} & 62.97 & \textbf{89.67} & \textbf{82.63} \\
        
        \midrule
        CLIP & 88.61 & 68.52 & 93.85 & 86.38 \\
        ComCLIP$^*$ & 89.84 & \textbf{69.92} & 93.61 & 86.93 \\
        \textbf{GCLIP}$_{\text{ours}}$ & \textbf{92.38} & 69.45 & \textbf{94.27} & \textbf{87.85} \\

         \midrule
        BLIP2 & 96.48 & 83.52 & 95.63 & 93.00 \\
        ComBLIP2$^*$ & 96.28 & \textbf{82.81}& 95.39 & 92.66 \\
        \textbf{GBLIP2}$_{\text{ours}}$ & \textbf{97.14} & 82.73 & \textbf{95.94} & \textbf{93.15} \\
        \bottomrule
    \end{tabular}
    \caption{Image-text matching accuracy of different models and approaches on the ComVG dataset. ($^*$\textit{The numbers are reproduced by us.})}
    \label{tab:comvg_results}
\vspace{-7pt}
\end{table}

\noindent \textbf{Evaluation Metric:} We use accuracy as the evaluation metric for both the ComVG and SVO-Probes. Each test example consists of a caption paired with one positive image and one negative image. A prediction is considered correct if the similarity score between the caption and the positive image is higher than that of the negative image. The final accuracy is computed as the percentage of correct predictions across the dataset. In addition to overall accuracy, we report accuracy separately based on which component (subject, predicate, or object) differs between the two images. 
\subsubsection{Results on ComVG and SVO-Probes}
Table~\ref{tab:comvg_results_1} and ~\ref{tab:svo_results_1} present image-text matching accuracy on ComVG and SVO Probes across different vision encoder backbones: ResNet-50, ViT-B/32, and ViT-L/14. Across all baselines: CLIP, OpenCLIP, ComCLIP and our proposed GCLIP, we observe that ViT-L/14 consistently achieves the highest accuracy, demonstrating stronger visual representations for compositional tasks. Based on this observation, we use ViT-L/14 as the default vision encoder in all subsequent evaluations.\\

\noindent Table~\ref{tab:comvg_results} and ~\ref{tab:svo_results} provide a more detailed breakdown of performance across compositional categories: subject, relation, object. These results offer direct evidence in support of our central claim that explicit grounding of entities enables more accurate  visual-textual alignment. Note that CLIP embeddings continue to struggle with verb disambiguation in {\bf Rel} column and BLIP2 does notably better in all columns suggesting the effectiveness of an additional training stage when compared to CLIP. Examples of correct and incorrect matches can be found in Figures~\ref{fig:holding}, and \ref{fig:swinging}.
\\

\begin{table}[h]
    \centering
    \begin{tabular}{lcccc}
        \toprule
        \textbf{Method} & \textbf{Sub} & \textbf{Rel} & \textbf{Obj} & \textbf{Total} \\
        \midrule
        SLIP & 81.71 & 72.80 & 89.14 & 77.59 \\
        ComSLIP$^*$ & 82.73 & 73.39 & 90.38 &  78.39\\
        \textbf{GSLIP}$_{\text{ours}}$ & \textbf{86.54} & \textbf{73.76} & \textbf{91.90} & \textbf{80.09} \\

        \midrule
        CLIP & 87.46 & 82.31 & 90.27 & 84.79 \\
        ComCLIP$^*$ & 88.34 & 82.74 & 90.90  & 85.32  \\
        \textbf{GCLIP}$_{\text{ours}}$ & \textbf{89.54} & \textbf{83.27} & \textbf{91.75} & \textbf{86.29} \\
        \midrule
        BLIP & 83.13 & 82.54 & 89.14 & 83.99 \\
        ComBLIP$^*$ & 85.09 & 83.63 & 90.41 & 85.25  \\
        \textbf{GBLIP}$_{\text{ours}}$ & \textbf{85.49} & \textbf{84.21} & \textbf{90.52} & \textbf{85.85} \\

        \midrule
        BLIP2 & 92.75 & 89.46 & 96.27 &  91.39\\
        ComBLIP2$^*$ & 92.97 & 89.18  & 96.70 & 91.33 \\
        \textbf{GBLIP2}$_{\text{ours}}$ & \textbf{93.78} & \textbf{89.87} & \textbf{97.36} & \textbf{92.24} \\
        \bottomrule
    \end{tabular}
    \caption{Image-text matching accuracy of different models and approaches on the SVO-Probes dataset.($^*$\textit{The numbers are reproduced by us.})}
    \label{tab:svo_results}
\vspace{-7pt}
\end{table}

\noindent In Table~\ref{tab:comvg_results} on the ComVG dataset, GCLIP consistently improves performance across all components. In the subject category, GCLIP outperformed ComCLIP (92.38\% vs. 89.84\%) and CLIP (92.38\% vs. 88.61\%). This is crucial because, dataset biases often affect subject recognition
 and models struggle when the same objects appear in different positions across training data. By localizing and comparing grounded embeddings, our method more reliably distinguishes between semantically distinct configurations. The object category follows a similar trend, where GroundingCLIP reaches 95.94\%, surpassing all prior methods. This suggests that grounding improves not only fine-grained subject recognition but also reinforces object-level distinctions in scenes with visually similar elements. \\

\noindent In Table~\ref{tab:svo_results}, on the SVO-Probes dataset, which features more diverse and unconstrained image-text pairs, GCLIP outperformed both CLIP and ComCLIP across all categories. This suggests that the benefits of grounding are not just limited to structured or synthetic datasets, but also generalize well to real-world compositions and small variations in phrasing and visual layout. \\

\noindent ComCLIP's minor advantage on ComVG relation accuracy likely comes from
GRiT captioning model which is partly being trained on Visual Genome dataset. SVO Probes images do not come from Visual Genome dataset and the verbs are more ambiguous. Hence, GCLIP's direct box-based relation grounding provides a more robust and accurate signal, resulting in better performance.


\subsection{Image Text Retrieval}
To evaluate the generalization ability of GCLIP beyond structured triplets, we have conducted retrieval experiments on Flickr30K and MS-COCO, two benchmarks featuring natural, free form human written captions. \\

\noindent Our retrieval pipeline begins by prompting GPT-3.5 to analyze the input caption in two stages:

\begin{itemize}
\item In the first stage, we prompt the model to identify all objects mentioned in the caption, extract their associated attributes if any (eg: "red shirt") and describe how these objects are connected through subject-verb-object relations. 
\item In the second stage, we again prompt the same model to combine the objects with their attributes and generate short natural phrases. These noun phrases are then passed to GroundingDINO to detect and localize the corresponding visual regions in the image. 
\end{itemize}

\noindent For each caption, we apply this process to the top 10 retrieved images. Each image is treated as an individual image-text matching instance, where we use the localized bounding boxes and compositional fusion (Section \ref{sec:image_text_matching}) to compute a refined Image Text Matching score (the cosine similarity between the enriched image embedding and the full caption embedding.). These scores are then used to re-rank the 10 candidates, improving the precision of final retrieval.\\

\noindent 
An added advantage of our approach is the use of global image embedding in combination with the grounded region embeddings. The global embedding offers contextual information that may not be captured by a single region, while the grounded sub images allow the model to focus on specific entities and relations mentioned in the caption. This is very crucial in retrieval where the free form captions can contain overall scene details such as background or scene context that span beyond individual objects.
Using the weighted sum of grounded region embeddings and the global image embedding, our method achieves a balance between scene-level semantics and fine-grained compositional alignment. \\

\begin{figure}
\includegraphics[width=\columnwidth]{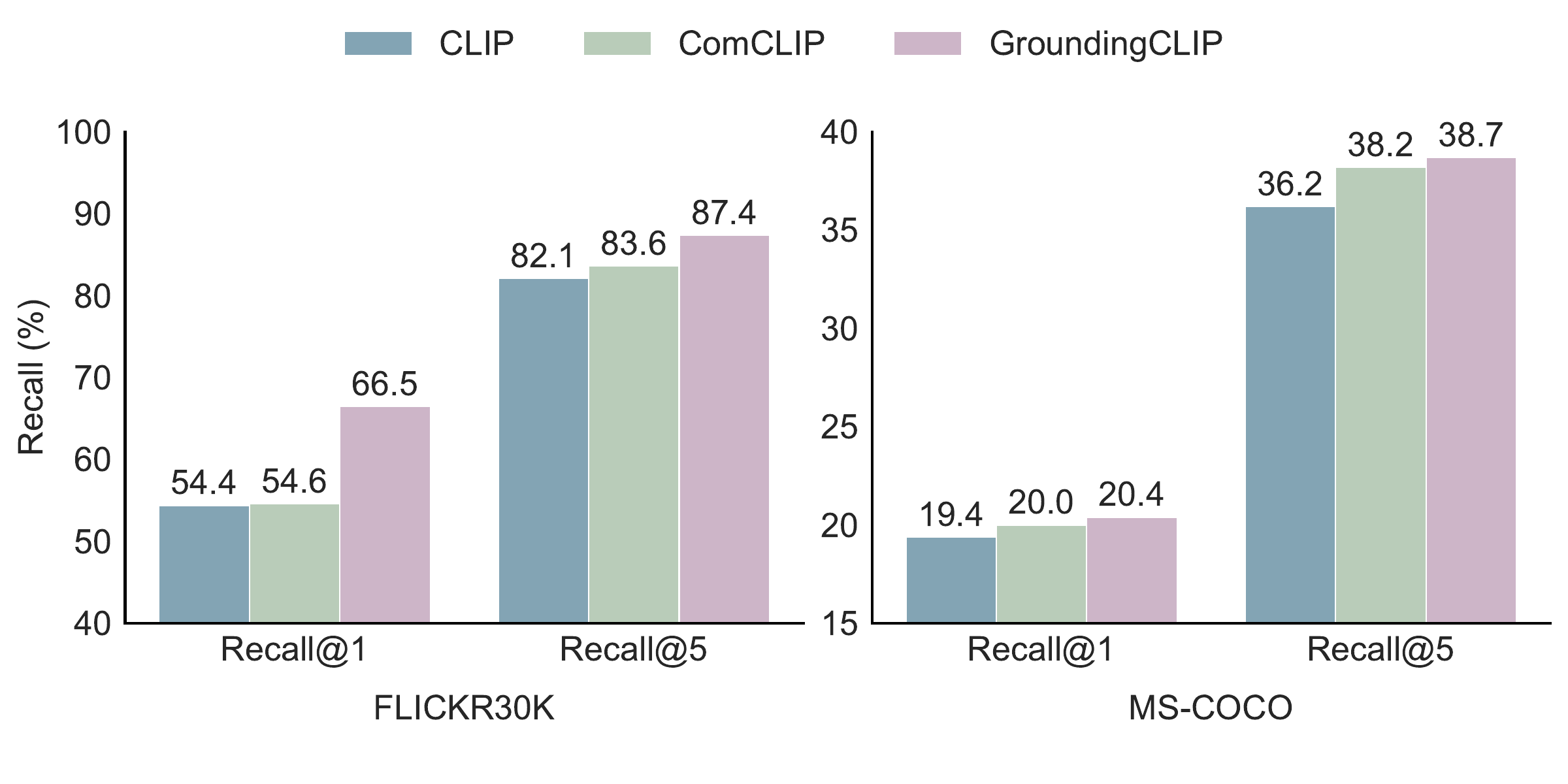} 
        \caption{Comparison of Recall@1 (\%) and Recall@5 on Flickr30K and MS-COCO datasets}   
        \label{fig:retrieval}
\vspace{-5mm}
\end{figure}

\noindent The results on Flickr30K and MS-COCO benchmarks from Figure~\ref{fig:retrieval} reveal significant improvement introduced by GCLIP and how it aligns images with free form captions. Unlike traditional models that solely rely on global image representation, GCLIP integrates localized entity specific visual details guided by textual phrases enabling it to distinguish between visually similar scenes based on fine grained compositional details. A 12\% boost in Recall@1 on Flickr30K indicates a meaningful shift in retrieval behavior. These results underscore that retrieval performance can be significantly enhanced by redefining the image representation through grounding, without requiring any additional training.

\section{Conclusions}

We proposed Grounding CLIP (GCLIP), a training-free method for enhancing compositional image-text alignment by dynamically refining global image embeddings using grounded sub-regions detected via an open-vocabulary detector. Our method consistently outperformed CLIP and ComCLIP across both structured (ComVG, SVO Probes) and natural (Flickr30K, MS-COCO) benchmarks, achieving significant improvements in image-text matching and retrieval tasks. The results highlight the importance of grounding individual entities and relations in improving fine-grained visual-textual alignment. In our Future work, we plan to explore additional benchmarks, models and fine-tuning strategies to further improve the model's ability to distinguish between different verbs and relations and consider scalability aspects on large scale retrieval datasets. 

\section{Limitations}
While GCLIP achieves strong performance without additional training, it has a few limitations. First, the reliance on large language models and open-vocabulary detectors which introduces additional inference time overhead, which can impact scalability in real-time applications. Second, the quality of the final image embedding is sensitive to the accuracy of entity grounding; failure to detect or correctly localize entities can negatively affect similarity computation. Finally, although GCLIP significantly improves alignment for subjects and objects, the performance on verb (relation) disambiguation still lags behind, likely due to the complexity involved in identifying actions and interactions that span across multiple visual regions.






{
    \small
    \bibliographystyle{ieeenat_fullname}
    \bibliography{main}
}


\end{document}